\documentclass[twocolumn]{htl-author}

\DOI{2012.0357}

\begin{document}

\title{Augmenting Efficient Real-time Surgical Instrument Segmentation in Video with Point Tracking and Segment Anything}

\author{Zijian Wu$^{1}$, Adam Schmidt$^{1}$, Peter Kazanzides$^{2}$ and Septimiu E. Salcudean$^{1}$}

\address{$^{1}$Robotics and Control Laboratory, Department of Electrical and Computer Engineering, The University of British Columbia,  Vancouver, BC V6T 1Z4, Canada\\
E-mail: zijianwu@ece.ubc.ca\\
$^{2}$Department of Computer Science, Johns Hopkins University, Baltimore, MD 21218, USA}


\abstract{The Segment Anything Model (SAM) is a powerful vision foundation model that is revolutionizing the traditional paradigm of segmentation.
Despite this, a reliance on prompting each frame and large computational cost limit its usage in robotically assisted surgery.
Applications, such as augmented reality guidance, require little user intervention along with efficient inference to be usable clinically.
In this study, we address these limitations by adopting lightweight SAM variants to meet the efficiency requirement and employing fine-tuning techniques to enhance their generalization in surgical scenes.
Recent advancements in Tracking Any Point (TAP) have shown promising results in both accuracy and efficiency, particularly when points are occluded or leave the field of view.
Inspired by this progress, we present a novel framework that combines an online point tracker with a lightweight SAM model that is fine-tuned for surgical instrument segmentation.
Sparse points within the region of interest are tracked and used to prompt SAM throughout the video sequence, providing temporal consistency.
The quantitative results surpass the state-of-the-art semi-supervised video object segmentation method XMem on the EndoVis 2015 dataset with 84.8 IoU and 91.0 Dice. Our method achieves promising performance that is comparable to XMem and transformer-based fully supervised segmentation methods on \textit{ex vivo} UCL dVRK and \textit{in vivo} CholecSeg8k datasets. In addition, the proposed method shows promising zero-shot generalization ability on the label-free STIR dataset. In terms of efficiency, we tested our method on a single GeForce RTX 4060/4090 GPU respectively, achieving an over 25/90 FPS inference speed. Code is available at: https://github.com/wuzijian1997/SIS-PT-SAM}
\maketitle

\section{Introduction}
Surgical instrument segmentation (SIS) is a fundamental task that provides essential visual cues for various downstream applications of robotic surgery, including augmented reality \cite{ref_article1} and surgical scene understanding \cite{ref_article2,ref_article3}. 
Segmenting surgical tools from the tissue background is challenging due to occlusion, blood, smoke, motion artifacts, and changing illumination.
While deep learning-based segmentation methods have made significant strides in recent years, achieving high accuracy relies on training using large-scale datasets with annotated images.
In surgical computer vision, high-quality annotation is particularly scarce due to the time-consuming, labor-intensive, and expertise-demanding process of labeling. 

Recently, the Segment Anything Model (SAM)~\cite{ref_article5}, the first promptable foundation model for image segmentation, has attracted widespread attention.
SAM's demonstrated impressive zero-shot generalization capability along with its flexible prompting framework make it especially useful for enabling downstream applications.
In surgical scenarios, however, the application of SAM faces two challenges.
Firstly, the huge computational cost of its heavyweight image encoder architecture, especially when processing high-resolution images, hinders its real-time inference capabilities~\cite{ref_article6}. 
Furthermore, numerous studies have reported significant performance degradation of SAM on medical images~\cite{ref_article15,ref_article16}, including images in surgical scenes~\cite{ref_article14}.
In this study, we adopt the lightweight SAM variant to facilitate inference efficiency.
Furthermore, we investigate the point prompt-based fine-tuning strategy for MobileSAM~\cite{ref_article7} to mitigate the performance degradation associated with the lightweight network architecture.

Despite SAM's strong automatic mask generation ability, achieving expected segmentation results in practice often requires appropriate prompts. 
Providing specific points or descriptive text of the target object can significantly improve the segmentation accuracy.
Leveraging the long-term tracking capabilities of the Tracking Any Point (TAP) models, we employ an online point tracker, CoTracker~\cite{ref_article9}, to provide sparse point prompts for SAM. Similar to SAM-PT~\cite{ref_article8} and DEVA~\cite{ref_article20}, our pipeline decouples video object segmentation (VOS) into image-level segmentation, which can be task-specific, and a universal temporal propagation.
Compared to end-to-end VOS, our ``tracking-by-detection" framework can take full advantage of smaller image-level datasets via fine-tuning a task-specific image segmentation model and using it in tandem with a point tracker to maintain temporal consistency.   

In summary, our contribution is twofold: (1) we present a real-time video surgical instrument segmentation framework that achieves superior segmentation performance and is suitable for clinical usage due to its good efficiency; (2)
we investigate the point prompt-based fine-tuning strategy (will open source) for lightweight SAM using surgical datasets, and the model fine-tuned on only two datasets shows promising generalization on unseen surgical videos.

\section{Related work}
The goal of Tracking Any Point (TAP) is to estimate the motion of arbitrary physical points throughout a video.
TAP-Vid~\cite{ref_lncs4} first formalized this task alongside a benchmark dataset and baseline method for TAP.
Recent work has showcased the promising online TAP capability and exhibited great robustness to occlusion and exit from the field of view.
PIPs++~\cite{ref_lncs5} and TAPIR~\cite{ref_article10} demonstrate substantial robustness under occlusion and achieve real-time inference speed on high-resolution video.
Notably, CoTracker~\cite{ref_article9} achieves state-of-the-art tracking performance by jointly tracking a set of query points.
CoTracker is an online algorithm that processes video sequentially through a sliding window.
Optical flow~\cite{ref_lncs2,ref_lncs3} can be used for TAP but tends to accumulate errors over time and faces challenges in handling occlusions, which are common occurrences in surgical scenarios.
 
SAM is the first vision foundation model for image segmentation, trained over the SA-1B dataset consisting of 1 billion high-quality annotated images~\cite{ref_article5}.
SAM demonstrates impressive zero-shot inference capability on natural images and supports flexible prompts.
Nevertheless, SAM's performance often declines in specific fields~\cite{ref_article14,ref_article15,ref_article16,ref_article19}, which can be attributed to a substantial domain gap.
Much research has been dedicated to adapting SAM to medical images~\cite{ref_article11,ref_article13} including surgical images~\cite{ref_lncs6,ref_article12}.
SurgicalSAM~\cite{ref_lncs6} and AdaptiveSAM~\cite{ref_article12} are adapted to the surgical domain by providing class and text prompts.
However, neither has real-time inference speed. 
The computational cost of SAM stems from its heavy image encoder, with some research~\cite{ref_article6,ref_lncs1,ref_article7,ref_article25} aiming to accelerate SAM's inference and reduce the demand for computation resources.
\section{Method}
Our proposed framework consists of two key components: a point tracker and a point-based segmentation model. 
Both components can be flexibly replaced with state-of-the-art models. 
As shown in Fig. 1, the pipeline can be described as follows. 
To begin with, the first frame mask of the video sequence is generated to indicate the region of interest (ROI) where query points are initialized. 
Subsequently, a set of query points is selected within the ROI based on a sampling strategy.
After this pre-processing, we employ a point tracker to track these query points and utilize them as prompts at each frame for the segmentation model.

In Section 3.1, we illustrate the pre-processing that is used to initialize query points. 
In Section 3.2, we formalize and clarify the proposed TAP + SAM framework.
In Section 3.3, we introduce the SAM fine-tuning strategy employed in this study.
\begin{figure}[!h]\label{fig1}
\centering{\includegraphics[width=8cm]{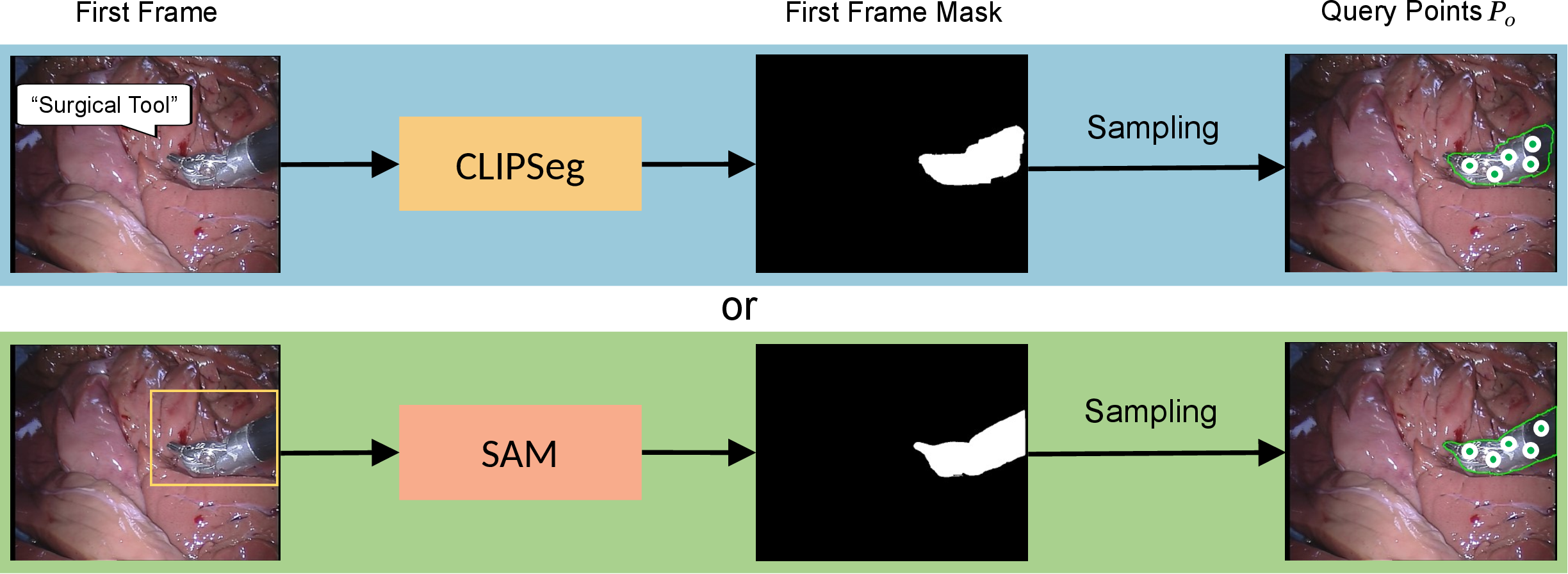}}
\caption{The pre-processing workflow to generate the query points. The segmentation model can be semi-automatic, i.e., SAM + bounding box prompt (bottom row), or fully automatic, i.e., CLIPSeg + fixed text prompt “surgical tool” (top row). Note that, without the initial mask, just manually picking query points is also feasible.}
\end{figure}
\subsection{Pre-processing}
We use the SAM model to generate the initial mask by simply inputting a few points or bounding boxes. 
Inspired by recent advancements in vision-language models, several zero-shot segmentation models based on text prompts have been developed~\cite{ref_lncs9,ref_lncs10}.
To achieve the ``fully" automatic pipeline, we incorporate a text-promptable segmentation model CLIPSeg~\cite{ref_lncs9} to automatically generate the initial mask by setting a text prompt, ``surgical tool". 
While CLIPSeg can only provide a coarse initial mask, it remains feasible for query point selection as long as its output roughly covers the region of the target instrument.

We initialize query points using the first frame mask. 
We investigate various query point sampling strategies, including random sampling, uniform sampling on a grid, SIFT keypoints, Shi-Tomasi corner points, and K-Medoids~\cite{ref_book1} clustering centers.
We selected K-Medoids clustering centers because they ensure even partitioning of the entire cluster.
The number of medoids assigned to each instance ranges from 1 to 9, and we choose 5 in the experiments.
Let
\begin{equation}\label{equation4}
P_0 = \{(p_i, t_0)\}, \,\,\,\,  p_i = (x_i, y_i), \,\, i=1,...,N
\end{equation}
be the initial set of query point locations $p_i$ at time $t_0$.
\subsection{Tracking Any Point + Segment Anything}
\begin{figure}[!h]
\centering{\includegraphics[width=8cm]{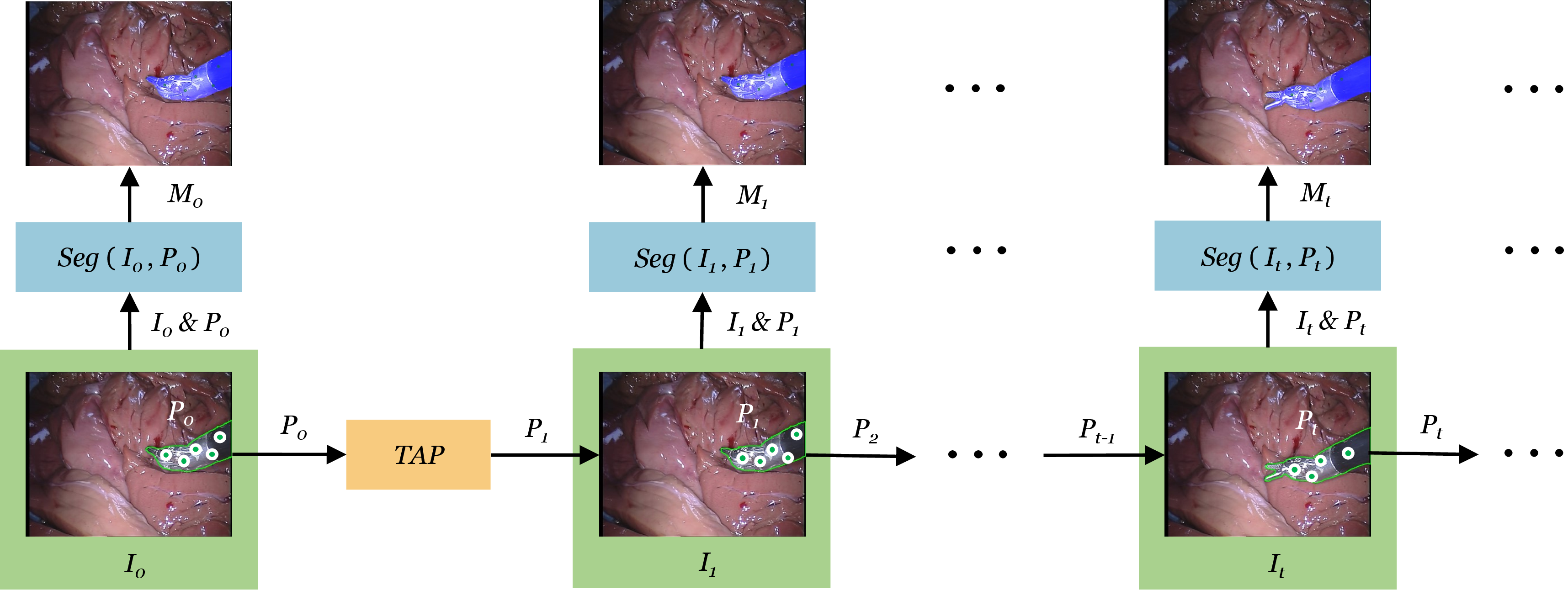}}
\caption{The overview of our video surgical instrument segmentation pipeline that combines a segmentation model $Seg(\cdot)$ and a point tracker $TAP(\cdot)$.}
\end{figure}
We describe our framework with reference to Fig.~2. 
Given a video $V = \{I_t\},$ in which $I_t\in R^2$ is the image at time $t$, along with a set of initial query points $P_0$ in $I_0$, we use TAP to predict the query points $P_t=\{(p_i,\;t)\}$, at time $t$
\begin{equation}
P_t=TAP(V,\;P_0),
\end{equation}
and we use $P_t$ and the current image $I_t$ in a segmentation model
\begin{equation}
    M_t=Seg(I_t, P_t). 
\end{equation}
to produce the mask $M_t$.

We incorporate the state-of-the-art model, CoTracker~\cite{ref_article9}, as the off-the-shelf online point tracker to propagate the initial query points throughout the video sequence.
CoTracker takes a short video clip consisting of several frames as the input. 
It processes video frames in a serial fashion via a 4-frame sliding window, making it suitable for online applications. 
We also integrated PIPs++~\cite{ref_lncs5} and TAPIR~\cite{ref_article10} into our software, but only CoTracker is used for experiments in this paper. 
For the segmentation model $Seg$, we adopt the fine-tuned MobileSAM to enable real-time processing throughout the entire pipeline, while achieving accurate segmentation.  
\subsection{Fine-tuning the Segment Anything Model}
As for the segmentation model, we first tested two state-of-the-art lightweight SAM variants, MobileSAM and Light HQ-SAM~\cite{ref_lncs1}.
However, we observed that both methods perform poorly in situations where specularity, blood, or weak lighting is present, as shown in Fig.~3.
Driven by this limitation, enhancing the generalization of the lightweight SAM for surgical instrument segmentation becomes crucial. 
The state-of-the-art MedSAM~\cite{ref_article11} has demonstrated that fully fine-tuning SAM for medical images can yield promising results. Fully fine-tuning refers to freezing the prompt encoder and updating both the image encoder and mask decoder. Compared with strategies that only update the mask decoder or introduce an adapter layer, fully fine-tuning achieves superior performance.
\begin{figure}[!h]
\centering{\includegraphics[width=8cm]{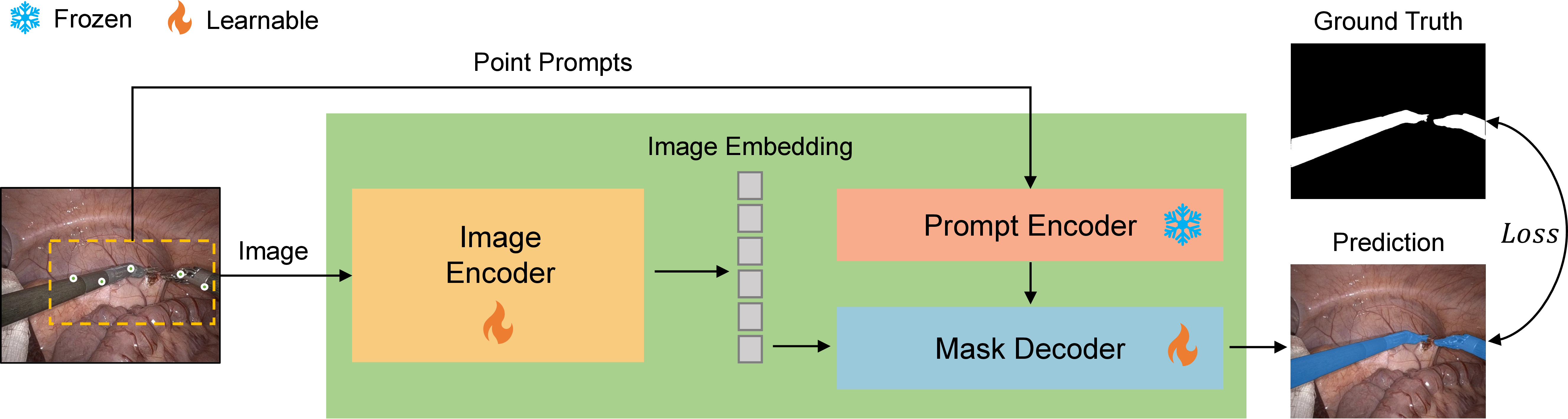}}
\caption{The pipeline of SAM fine-tuning using points. The input for the SAM model consists of images paired with points within the target area. The green rectangle represents the MobileSAM architecture.}
\end{figure}
In this study, we investigate the fully fine-tuning strategy for the most widely used lightweight SAM variant, MobileSAM, to optimize its capability for surgical instrument segmentation. As depicted in Fig.~3, the network architecture of MobileSAM is consistent with the original SAM. Unlike using bounding box prompts in MedSAM, we utilize point prompts during training to maintain consistency with the prompt type used during inference. Both the image encoder and mask decoder are learnable and updated during the fine-tuning. 
Due to the small number of learnable parameters of MobileSAM (10.13M), the cost of computation is significantly reduced.
We train our model on 4 V100 GPUs for 50 epochs. The training is based on the pre-trained MobileSAM weight. In datasets with instance-level labels, we randomly sample 5 points as prompts within the area of each instance. For datasets with binary labels, we randomly sample 5 points as prompts within the segmented region. The loss $L$ utilizes an unweighted combination of binary cross entropy loss and Dice loss~\cite{ref_lncs11}, represented as
\begin{equation}
L\;=\;L_{BCE}\;+\;L_{Dice}
\end{equation}
We used the AdamW~\cite{ref_article24} optimizer for training, with a batch size of 32. The initial learning rate is set to 1e-5 and follows a cosine decay schedule. All images are resized to 1024 by 1024 and undergo random up-down and left-right flip data augmentation. All images are Min-Max normalized and standardized.

\section{Results}
\subsection{Datasets}
We conduct quantitative comparisons on the EndoVis 2015~\cite{ref_article23}, UCL dVRK~\cite{ref_lncs12} and CholecSeg8k \cite{ref_article26} datasets to validate the feasibility and performance of our proposed framework. We finetune the SAM model using the training set of the ROBUST-MIS 2019 \cite{ref_article27} dataset and provide qualitative results because there is no video-level annotation for quantitative evaluation. 
To test the zero-shot generalization, we test our method on the unlabeled STIR~\cite{ref_article22} dataset and display qualitative results.

The EndoVis 2015 dataset provides 25 FPS videos and corresponding articulated da Vinci robotic instruments in \textit{ex vivo} background. It consists of four 45-second training videos and six testing videos (four 15-second and two 60-second videos). Note that there is only one type of instrument (needle driver) in the training set while two types (needle driver and scissor) in the testing set. We keep the original data split for fine-tuning and testing. 

The \textit{ex vivo} UCL dVRK dataset consists of 14 videos of 300 frames with corresponding binary segmentation masks. The dataset is split into training (Video 1-8), validation (Video 9 and 10) and testing (Video 11-14) sets. The videos are collected and annotated at 6.7~FPS. 

Based on the Cholec80 dataset \cite{ref_article28}, the CholecSeg8k dataset consists of 8080 frames (from 17 \textit{in vivo} cholecystectomy video clips) with segmentation ground truth. We choose 8 consecutive clips, which are 905 frames in total, as the testing set. The rest of the data is split into the training and validation set as 80\% and 20\% ratio. 

We select two extra datasets for qualitative evaluation. The ROBUST-MIS 2019 data comprises 10,040 annotated images, of which 5,983 are in the training set. The testing set is divided into three stages with increasing levels of difficulty. The dataset provides one labeled frame per clip. The STIR dataset consists of high-resolution label-free videos collected by da Vinci Xi.
\subsection{Quantitative Results} 
We perform comparisons between the state-of-the-art semi-supervised VOS method XMem~\cite{ref_lncs13}, and fully supervised image segmentation methods TransUNet \cite{ref_article18} and SwinUNet \cite{ref_lncs15}. The ablation study of different SAM variants (MobileSAM, HQ-SAM Light, and the default SAM) demonstrates the significant performance improvement of finetuning. All these SAMs use the CoTracker as the online point tracker.
We adopt the widely-used segmentation metrics, $IoU$ and $Dice$, for quantitative comparison and ablation study. 
Table 1 displays the quantitative results on three datasets, in which the first 3 rows are comparison results, the next 3 rows are the ablation results, and the bottom row is ours. 
\begin{center}
\begin{table}[!h]
\caption{Quantitative Results of Different Datasets}
\resizebox{\linewidth}{!}{
\begin{tabular}{lcccccc}
  \hline
  Methods & \multicolumn{2}{c}{EndoVis 2015 \cite{ref_article23}} & \multicolumn{2}{c}{UCL dVRK \cite{ref_lncs5}} & \multicolumn{2}{c}{CholecSeg8k \cite{ref_article26}} \\ \cline{2-7}
  & IoU & Dice & IoU & Dice & IoU & Dice \\
  \hline
  TransUNet \cite{ref_article18} & 57.7 & 71.8 & 79.1 & 88.0 & 71.7 & 82.8\\
  SwinUNet \cite{ref_lncs15} & 59.5 & 73.0 & 81.1 & 89.3 & 81.5 & \textbf{89.5} \\
    XMem \cite{ref_lncs13} & 82.6 & 89.3 & \textbf{91.9} & \textbf{95.4} & 81.6 & 87.9 \\
  \cline{1-7}
  PT+MobileSAM \cite{ref_article7} & 70.1 & 80.6 & 45.0 & 57.1 & 49.7 & 57.4 \\
  PT+HQ-SAM Light \cite{ref_lncs1} & 69.0 & 80.2 & 55.9 & 68.1 & 61.3 & 70.4 \\
  PT+SAM (ViT-H) \cite{ref_article5} & 79.6 & 88.3 & 74.3 & 83.3 & 64.5 & 70.6\\
  \cline{1-7}
  \textbf{Ours} & \textbf{84.4} & \textbf{91.0} & 89.4 & 93.8 & \textbf{81.9} & 88.6 \\
  \hline
\end{tabular}}
\end{table}
\end{center}

\begin{figure}[!h]\label{fig4}
\centering{\includegraphics[width=8cm]{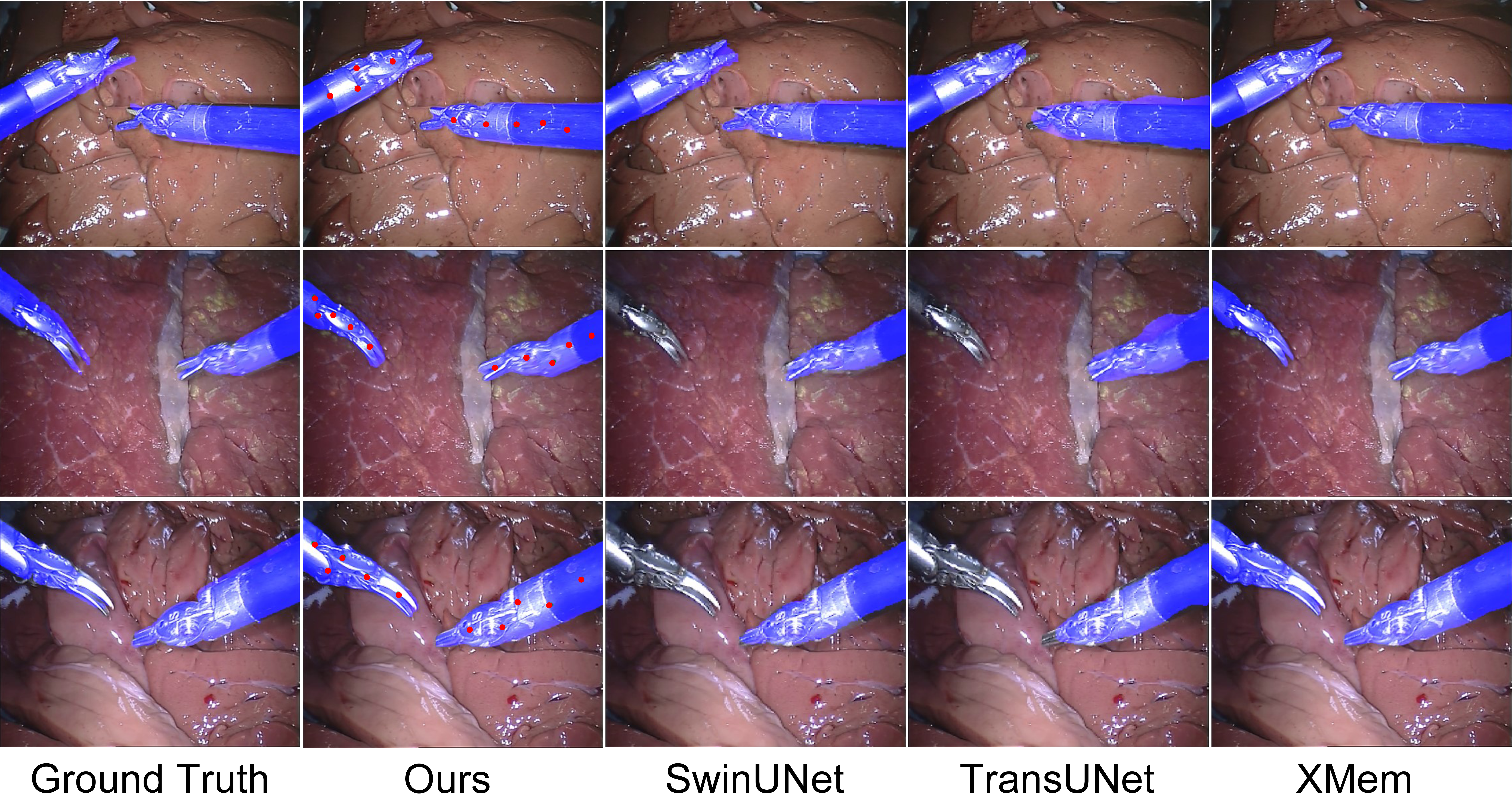}}
\caption{Visualization of segmentation results from several methods on the EndoVis 2015 dataset, in which the images are acquired from the testing Video 1, 5, and 6, respectively (from top row to bottom row). Note that red dots in Fig. 4 - 8 are the point prompts tracked by CoTracker.}
\end{figure}

\begin{figure}[!h]\label{fig5}
\centering{\includegraphics[width=8cm]{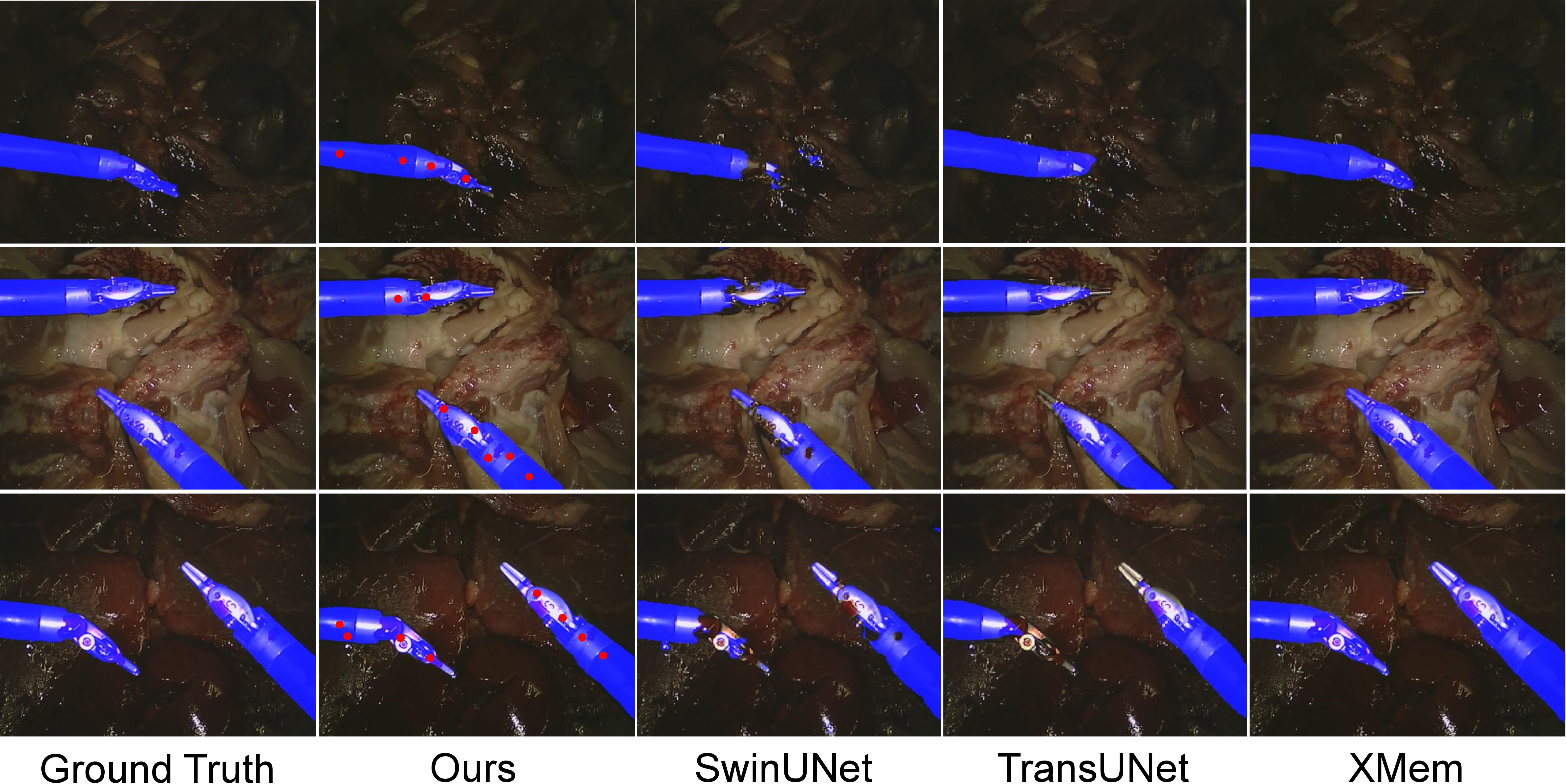}}
\caption{Visualization of segmentation results from several methods on the UCL dVRK dataset, in which the images are acquired from the testing Video 1, 3, and 4, respectively (from top row to bottom row). }
\end{figure}
\begin{figure}[!h]\label{fig6}
\centering{\includegraphics[width=8cm]{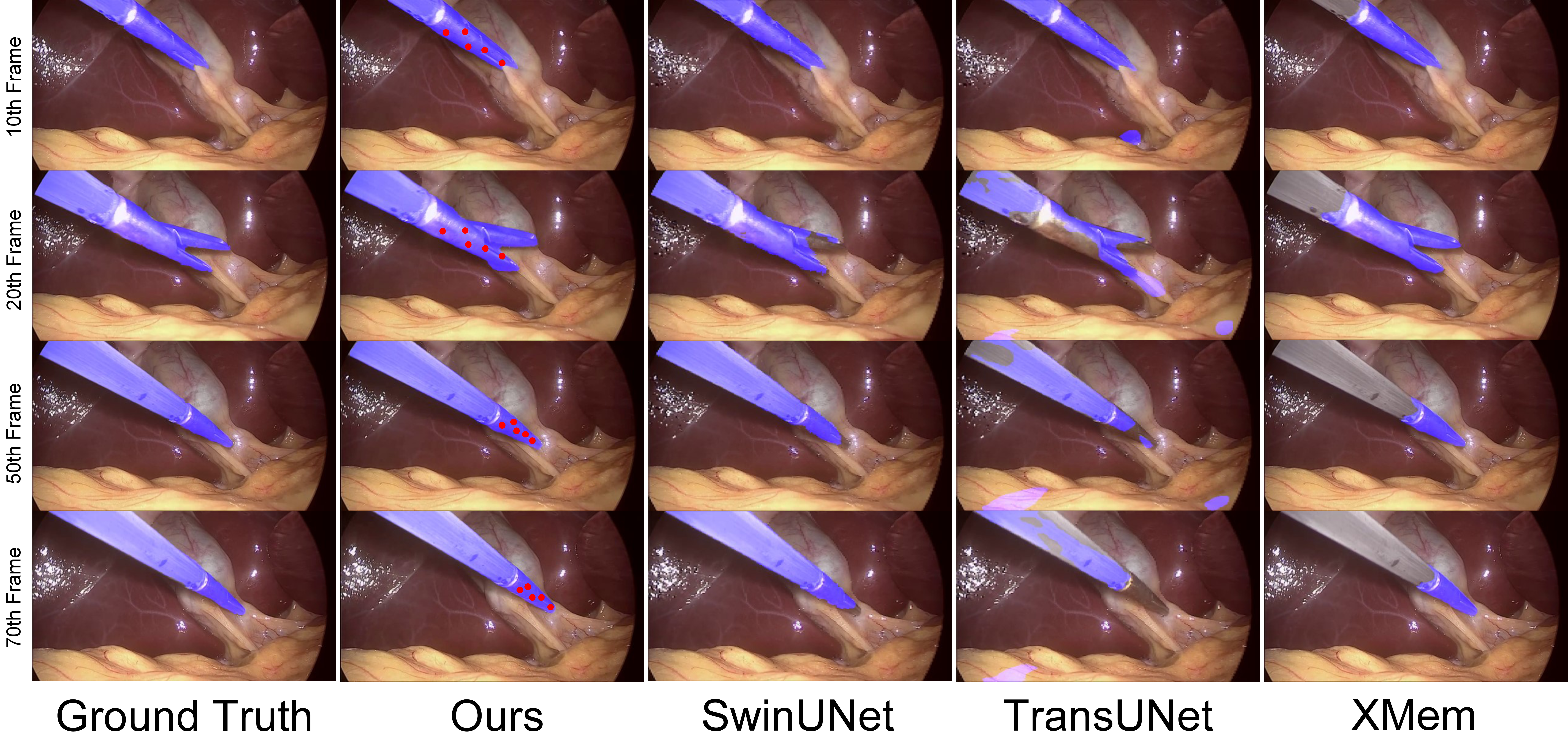}}
\caption{Segmentation results of different methods on the CholecSeg8k dataset. The y-axis is along the frame order (10th, 20th, 50th, and 70th) in one clip.}
\end{figure}
\begin{figure}[!t]\label{fig7}
\centering{\includegraphics[width=8cm]{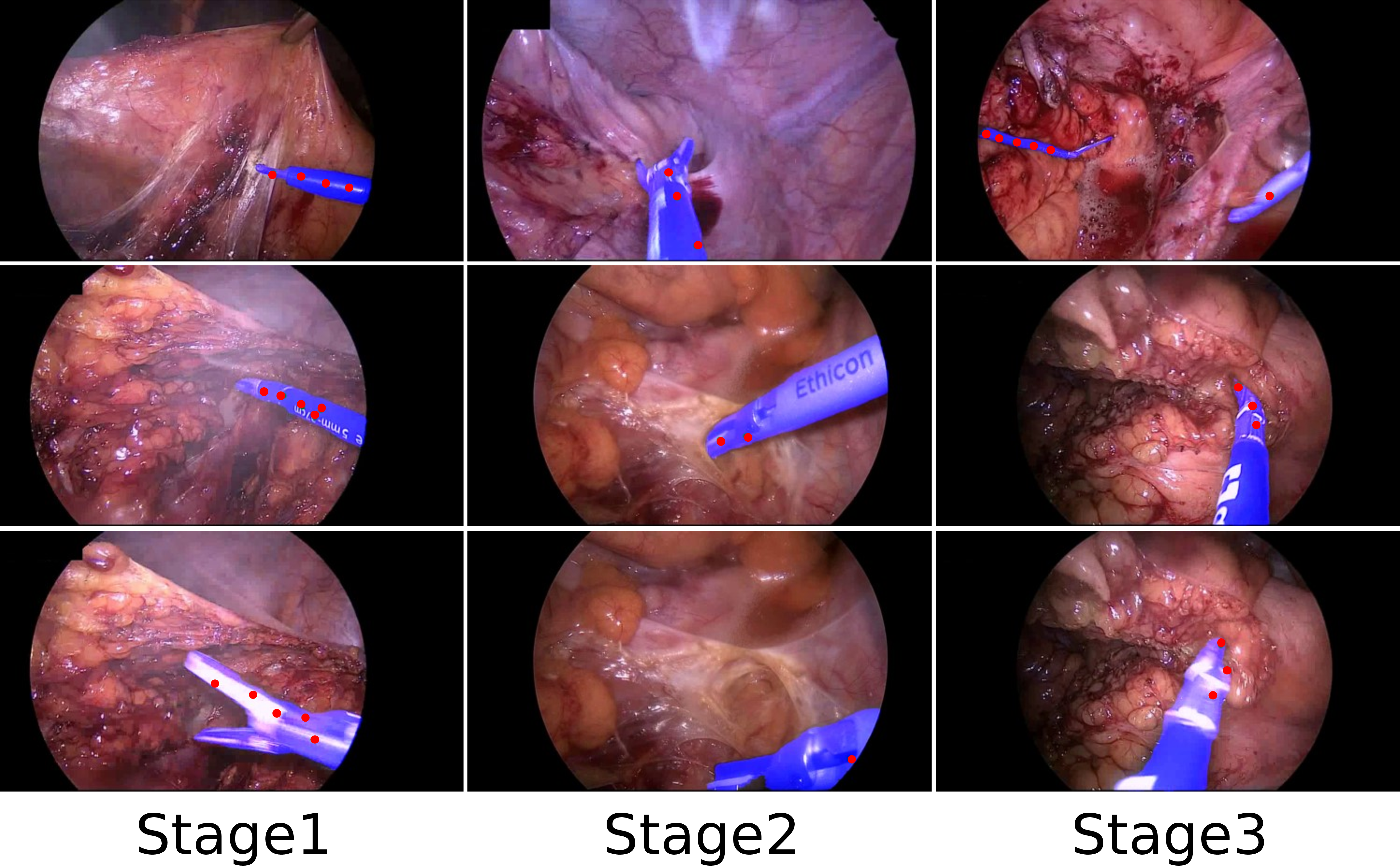}}
\caption{Qualitative results from our proposed methods on the ROBUST-MIS 2019 dataset, in which columns from left to right are the samples from stages 1, 2, and 3, respectively.}
\end{figure}

\subsection{Qualitative Results} 
Fig. 4 and Fig. 5 display a few predicted masks from various methods on EndoVis 2015 and UCL dVRK datasets. Fig. 6 is the segmentation results of the frame 10, 20, 50, and 70, which are sampled on a video clip from the testing set of the CholecSeg8k dataset. Fig. 7 is the visualization of segmentation results on different difficulty testing stages of the ROBUST-MIS 2019 dataset.
To evaluate the generalization of our method, we test our framework, which is fine-tuned using the EndoVis 15 and UCL dVRK dataset, on some videos from the STIR dataset.
Fig. 8 shows some segmentation results of our method, in which frame 397 is the end frame. The segmentation performance of our method is robust in this video.  
\begin{figure*}[!t]\label{fig8}
\centering{\includegraphics[width=17cm]{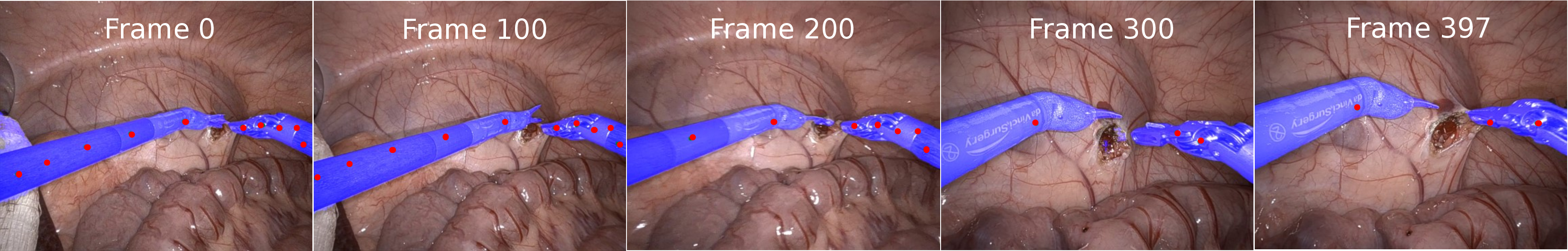}}
\caption{Visualization of qualitative results from our proposed methods on STIR dataset. }
\end{figure*}
\subsection{Efficiency} 
Table 2 displays the metrics of efficiency of different methods. The inference latency, inference memory, and learnable parameters represent the time efficiency, computation efficiency for inference, and computation efficiency for training. We run the inference procedure of each method on a laptop with RTX 4060 GPU and a desktop with RTX 4090 GPU. The inference latency of our method achieves 38 ms (26 FPS) and 11 ms (90 FPS), which are real-time and over real-time inference speed, on these two machines. The CoTracker's inference speed fluctuates in the 50-60~FPS range. In terms of computational efficiency, our proposed method only requires 2.8G inference memory. Our method is also training efficient with a small 10.1M parameters. We only list learnable parameters of partial methods, because the other methods do not involve training in this study.  
\begin{center}
\begin{table}[!h]
\caption{Efficiency Metrics with RTX 4090 and 4060 GPUs}
\centering
\resizebox{\linewidth}{!}{
\begin{tabular}{lcccc}
  \hline
  Methods & \multicolumn{2}{c}{Inference} & Inference  & Learnable \\
  & \multicolumn{2}{c}{Latency (ms)} & Memory (G) & Param. (M)\\
  \cline{2-3}
  & 4090 & 4060 &  & \\
  \hline
  XMem \cite{ref_lncs13} & 4 & 5 & 2.0 & -\\
  TransUNet \cite{ref_article18} & 7 & 8 & 2.8 & 105.3\\
  SwinUNet \cite{ref_lncs15} & 4 & 7 & 1.9 & 27.2  \\
  PT+MobileSAM \cite{ref_article7}& 11 & 38 & 2.8 & -\\
  PT+Light HQ-SAM \cite{ref_lncs1} & 12 & 40 &2.8 & - \\
  PT+ViT-H SAM \cite{ref_article5} & 220 & 1300 &8.2 & - \\
  \textbf{Ours} & 11 & 38 &2.8 & 10.1  \\
  \hline
\end{tabular}}
\end{table}
\end{center}

\section{Discussion} 
Overall, our method outperforms the state-of-the-art semi-supervised VOS model,  XMem, on the EndoVis 2015 dataset. Notably, in the one-minute videos, Video 5 and 6, our proposed method exhibits obvious improvement compared to XMem. This shows that our framework can robustly leverage the temporal information provided by universal TAP methods. The TransUNet and SwinUNet cannot recognize the scissors in testing videos because of the overfitting to the instrument type in the training videos.

As for the UCL dVRK dataset, our proposed method achieves promising performance with slightly weaker quantitative results than XMem.
All four UCL dVRK testing videos are captured under weak illumination conditions, which is unrealistic in real surgery. 
The dark scenes make it challenging for SAM to distinguish the boundary from the tissue background, especially when there are no such weak lighting conditions in the training set.
In contrast, the mask propagation-based XMem takes the full first frame ground truth for. 
The low video frame rate (6.7 FPS) of the UCL dVRK dataset also poses an obstacle for the TAP, thereby making it hard to provide effective point prompts throughout the video sequence.

The quantitative results on the \textit{in vivo} CholecSeg8k dataset achieve the best $IoU$ and second best $Dice$. The segmentation performance of our method is comparable to the state-of-the-art fully supervised image segmentation model SwinUNet.

The ablation study demonstrates the significance of fine-tuning by the substantial improvement compared to non-fine-tuned SAMs, even that of the powerful SAM with ViT-H backbone. Note that the inference speed of the ViT-H SAM is far away from real-time.

Compared to the propagation-based model such as XMem, our method has other advantages. XMem requires an accurate first frame ground truth for inference, while our pipeline only needs a known text prompt ``surgical tool".
Furthermore, XMem cannot recognize new objects during the video, while our method can tackle this by enabling users to easily pick a few new query points, instead of providing a high-quality mask.

\section{Limitations and Future Work} In general, our method achieves satisfied segmentation performance along with good efficiency in extensive SIS video datasets. However, when processing more challenging datasets like SAR-RARP \cite{ref_article29}, the segmentation performance of our method is not satisfactory, as depicted in Fig. 9. The videos of the SAR-RARP dataset are recorded during real robot-assisted laparoscopic prostatectomy with a cluttered scene, significant blood, specular reflection, wide-range rapid instrument motion, and frequent camera focus changes. These natures cause huge challenges for both point tracking and segment anything model. 

To enhance the performance in more challenging surgical scenarios, we plan to design a novel SAM prompt method to implicitly leverage the spatio-temporal information from point tracking instead of the naive combination. Kinematics data is a strong prior for instrument identification. However, this is only available for robotic surgical instruments. Other modality information such as text description of surgical instruments can be leveraged as a prior knowledge.  
\begin{figure}[!h]\label{fig9}
\centering{\includegraphics[width=8cm]{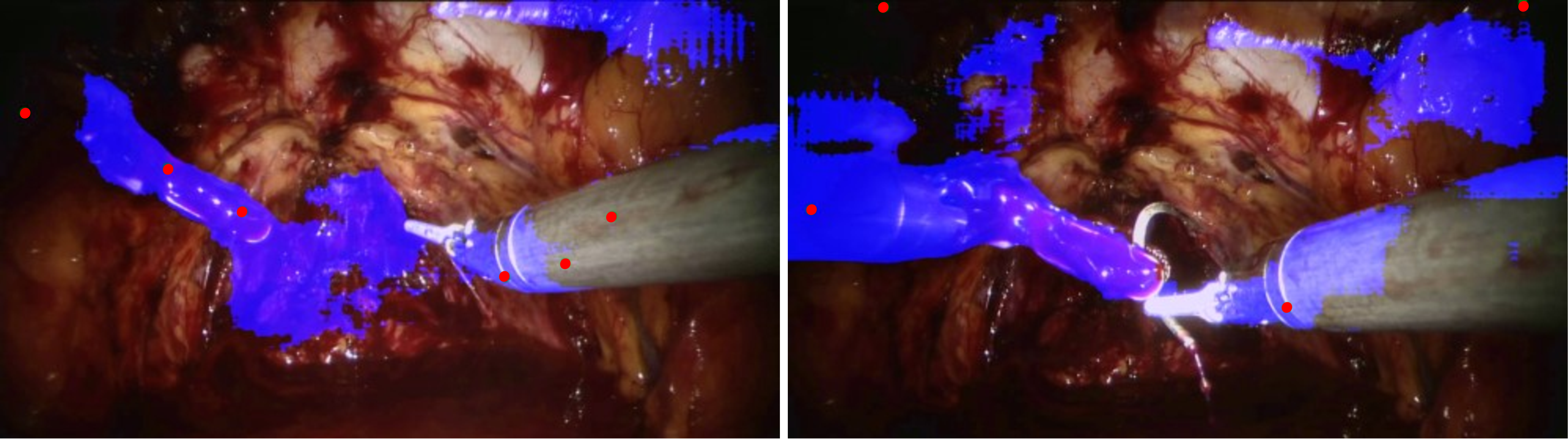}}
\caption{Failure cases on the SAR-RARP dataset.}
\end{figure}

\section{Conclusions}
In this study, we present a novel framework using a universal TAP and a fine-tuned lightweight SAM for real-time surgical instrument segmentation in video.  Its commendable efficiency and accuracy make it suitable for applications in clinical settings.
We investigate the availability of fine-tuning MobileSAM using point prompts and demonstrate the importance of fine-tuning for SIS. 
Extensive experiments validate the advancement of our proposed pipeline.
Furthermore, our SAM + TAP pipeline demonstrates the potential to serve as a strong VOS baseline by integrating other image segmentation models.



\end{document}